# Quantifying Gender Bias in Consumer Culture

Reihane Boghrati, Jonah Berger

Reihane Boghrati is an assistant professor of Information Systems at the W.P.Carey School of Business, Arizona State University (reihane.boghrati@asu.edu). Jonah Berger is an associate professor of Marketing at the Wharton School, University of Pennsylvania (corresponding author: jberger@wharton.upenn.edu).

# Quantifying Gender Bias in Consumer Culture

**ABSTRACT**

Cultural items like songs have an important impact in creating and reinforcing stereotypes, biases, and discrimination. But the actual nature of such items is often less transparent. Take songs, for example. Are lyrics biased against women? And how have any such biases changed over time? Natural language processing of a quarter of a million songs over 50 years quantifies misogyny. Women are less likely to be associated with desirable traits (i.e., competence), and while this bias has decreased, it persists. Ancillary analyses further suggest that song lyrics may help drive shifts in societal stereotypes towards women, and that lyrical shifts are driven by male artists (as female artists were less biased to begin with). Overall, these results shed light on cultural evolution, subtle measures of bias and discrimination, and how natural language processing and machine learning can provide deeper insight into stereotypes and cultural change.

*Keywords:* Natural language processing; discrimination; stereotypes; misogyny.

Consumer researchers have long been interested in stereotypes, bias, and discrimination (Brough et al. 2016; Douglas 1976; Maheswaran 1994). Research on gender bias, for example, finds that women are often perceived less favorably and treated less fairly across a range of situations (Carlana 2019; Garg et al. 2018; Mishra, Mishra, and Rathee 2019; Moss-Racusin et al. 2012; Reuben, Sapienza, and Zingales 2014). Targeting algorithms are more likely to show women search results and ads for products associated with negative attributes (Mishra, Rathee, and Mishra 2021) and the same job applicant is seen as more competent and offered a higher starting salary, if they have a male rather than a female name (Moss-Racusin et al. 2012).

One reason stereotypes, biases, and discrimination may be so sticky is that they are continually reinforced through culture. Songs, books, and other cultural items not only reflect the setting in which they were produced, but also shape the attitudes and behaviors of the audiences that consume them (Anderson, Carnagey, and Eubanks 2003; Berger et al. 2020; Lennings and Warburton 2011; Zayer et al. 2012). Song lyrics that are aggressive towards women, for example, or portray them negatively, increase anti-female attitudes and misogynous behavior (Fischer and Greitemeyer 2006). Lyrics that espouse equality, however, can boost attitudes towards women and encourage pro-female behavior (Greitemeyer, Hollingdale, and Traut-Mattausch 2015). Consequently, one reason stereotypes and biases, as well as attitudes more generally, may be so persistent is that they are continually reinforced by the cultural items (e.g., songs and advertisements) that consumers experience on an everyday basis.

But while such cultural items clearly have impact, their actual nature is less transparent. Consider music. Are song lyrics biased against women? Do any such biases vary across genres and artists? Have they changed over time?

Attempts to answer such questions have been hampered by issues of scale and measurement. While researchers in a range of disciplines have argued about whether music lyrics are misogynist (i.e., exhibit a dislike of, contempt for, or ingrained prejudice against women), most perspectives are based on small samples or one genre over a short period of time (Adams and Fuller 2006; Armstrong 2001; Flynn et al. 2016; Harding and Nett 1984; Herd 2009). Further, measuring misogyny is challenging. It's one thing to read and rate the lyrics of a few dozen songs, or even a few hundred, but rating the thousands of song lyrics necessary to truly get a scope of how misogyny has changed would be difficult. In addition, because existing analyses rely completely on human judgment, they are susceptible to bias. The same lyrics, for example, may seem more or less misogynistic depending on the perspective of the person reading them.

To address these challenges, we use natural language processing. We build a novel dataset of more than a quarter of a million songs from six music genres over more than 50 years. Then, using an emerging machine learning approach (i.e., word embeddings), and some simpler natural language processing techniques, we investigate potential biases and whether they have changed over time.

This work makes three main contributions. First, we showcase how an emerging machine learning method can shed light on bias, discrimination, and cultural change. While we focus on misogyny in music, the same approach could be used to study changes in how brands communicate with different audiences or talk about diversity, equity, and inclusion. We illustrate how the approach can be used, and then discuss broader applications.

Second, we deepen understanding around language and consumer behavior. Researchers have long been interested in this area (Lowrey and Shrum 2007; Lowrey, Shrum, and Dubitsky 2003), and recent work has begun to explore novel and exciting directions (Pogacar, Shrum, and Lowrey 2018), including gender (DeFranza, Mishra, and Mishra 2020; Mishra et al. 2019; Pogacar et al. 2021). We contribute to this emerging work, showcasing how natural language tools can deepen insight. In particular, we highlight the value of additional computational linguistic techniques. While more and more researchers are starting to use Linguistic Inquiry and Word Count (LIWC; Pennebaker et al. 2015) and similar dictionary-based methods, other approaches have received less attention. We discuss ways these approaches can shed light on a range of questions.

Third, we speak to the debate around misogyny in music. We provide the first large-scale analysis in this area, addressing (1) whether music is biased, (2) whether biases have changed over time, (3) what might be driving any changes, and (4) the potential impact of such changes (i.e., on attitudes towards women).

**LANGUAGE IN CONSUMER RESEARCH**

Recent work has begun to highlight the value of using automated textual analysis in consumer research (Mishra et al. 2019; Moore and McFerran 2017; Netzer et al. 2012; Netzer, Lemaire, and Herzenstein 2019; Packard, Moore, and McFerran 2018). Rather than requiring individuals to manually sift through texts, these computerized approaches extract a variety of textual features automatically. Researchers have used these approaches to measure things like gender bias in customer reviews (Mishra et al. 2019) and linguistic mimicry in online word of mouth (Moore and McFerran 2017). Automated textual analysis is particularly useful because it can parse large quantities of textual information, relatively quickly, and can do so in an objective manner (Chen, Nelson, and Hsu 2015; Zhang, Nrusimha, and Hsu 2018).

Most research in this area has relied on a class of approaches that can be described as word extraction. At a basic level, these approaches capture how often a given entity (e.g., word or phrase) appears in a text. Dictionaries like Linguistic Inquiry and Word Count (LIWC; Pennebaker et al. 2015), for example, allow researchers to measure how often first-person pronouns, positive emotion words, or other types of language show up. Tools like Evaluative Lexicon (Rocklage, Rucker, and Nordgren 2018), Hedonometer (Dodds and Danforth 2010), and VADER (Hutto and Gilbert 2014) use similar methods and researchers can even create their own customized dictionaries to use in a given application. But while word extraction is useful in a variety of applications, it has some limitations. Taken to the context of misogyny, for example, one could certainly measure the number of times certain pejorative terms for women show up in songs over time, but that, by itself, wouldn't tell you *how* those terms are being used. The word b*tch, for example, could be used to refer to either a man or a woman, or even an object.

Other work has begun to leverage a class of approaches called topic extraction. Similar to how factor analysis might be used to identify underlying themes among survey items, tools like topic modeling (e.g., Latent Dirichlet Allocation or LDA; Blei, Ng, and Jordan 2003) can identify the general topics or themes discussed in a body of text, as well as the words that make up those themes. Country songs, for example, talk a lot about the themes of "girls and cars" (e.g., words like car, drive, girl, and kiss) and "uncertain love" (e.g., words like love, ain't, and can't; Berger and Packard 2018). Toubia et al. (2019) use a similar approach to find the main themes discussed in movies.

But while topic extraction is useful for extracting themes from a large number of documents, it still doesn't say much about the relationship *between* entities. Whether women are talked about differently than men, for example, or whether certain brands are described as having particular traits.

For these types of questions, word embeddings is particularly useful (DeFranza et al. 2020; Devlin et al. 2018; Mikolov et al. 2013; Mishra et al. 2019; Pennington, Socher, and Manning 2014). As discussed in more detail in the methods, this approach represents each word by a vector, and uses a high-dimensional space to map each word or entity based on the other words with which it frequently appears. The relationship between these vectors captures the semantic relationship between words. Researchers can use this space to understand the relationship between words and the context in which they are used.

**THE CURRENT RESEARCH**

We use word embedding to examine whether women and men are talked about differently in music and whether any such biases may have shifted over time. We start by compiling a dataset of more than a quarter of a million songs from six music genres over more than 50 years, orders of magnitude larger than any previous dataset used to examine misogyny.

Next, we use several natural language processing approaches to examine potential biases. Before turning to the more complex word-embeddings approach, we begin with a simpler analysis. Since some work has measured misogyny using aggression towards women (Fischer and Greitemeyer 2006) we start by measuring the degree to which each gender is the recipient of aggressive thoughts and actions.

Gender bias can also be more subtle. Compared to men, for example, women are often described as warm (e.g., kind and supportive), but less likely to be described as competent (e.g., smart and ambitious; Fiske et al. 2002). Consequently, we use word embeddings to measure this subtler form of misogyny. Not whether lyrics are explicitly aggressive towards women, but whether woman are less likely to be linked with desirable traits.

Finally, we examine potential consequences and drivers and of any temporal shifts. We examine how such shifts are related to societal attitudes towards women and use artist gender to examine drivers. We examine whether the prevalence of female artists has shifted over time, whether male and female artists use different language, and how any shifts in their language are related to overall changes in misogyny.

We close with a broader discussion of how word embeddings, and natural language processing more generally, may be used to inform a wide range of research questions related to discrimination and marketing.

## EMPIRICAL ANALYSIS OF A QUARTER OF A MILLION SONGS

### *Data*

First, we compiled data on songs and their lyrics. Given copyright issues, there are limited open-access lyrics datasets, so we compiled information from different sources. We started by extracting all songs on each major Billboard chart (i.e., pop, rock, country, R&B, dance, and rap) every 3 months from 1965 (or whenever a given chart started) to 2018. This was then paired with lyrics from SongLyrics.com. To collect additional songs, we gathered all songs and their lyrics from datasets on kaggle.com (Agarwal 2017; Kuznetsov 2017; Mishra 2017) and scraped all available songs and their lyrics from a major lyrics website. Then, we used the million songs dataset (Mahieux et al. 2011) and freeDB (FreeDB 2018) to append year and genre for any song that did not already include this information. We removed any song with missing information and used artist and title combination to remove duplicates in the data.

The final dataset includes 258,937 songs from 1965 to 2018. It includes pop, rock, country, R&B, dance, electronic, rap, and hip-hop genres. Given data sparsity, and the similarity between rap and hip-hop, these two genres were combined. Dance and electronic were combined for the same reason.

Second, we cleaned the lyrics. Since we are focused on English lyrics, we used the langdetect package (version 1.0.7) in Python to remove any non-English lyrics. We also removed non-informative texts in brackets, such as [Verse 1] or [Pre-Chorus 1]. Finally, we shifted all lyrics to lower case so that variations of the same word are treated similarly (e.g., Man and man).

*Analyses*

**Measuring Aggression.** Prior work often measures misogyny using aggression towards women (Fischer and Greitemeyer 2006), so we begin by testing how frequently women are the object of aggression. While such analyses suggest that explicit aggression towards women has not changed (see Web Appendix), note that even misogynists might not explicitly sing about wanting to hurt women, for the fear that it would restrict their audience. More generally, if misogyny does exist it may be more implicit.

**Measuring Implicit Misogyny.** To address these limitations, we use a state-of-the-art machine learning approach to examine a subtler form of misogyny. Not whether lyrics are explicitly aggressive towards women, but whether woman are less likely to be linked with desirable traits (i.e., competence).

Competence and warmth are two universal dimensions of social cognition (Fiske, Cuddy, and Glick 2007), but while women are often described as warm (e.g., kind and supportive), they are less often described as competent (e.g., smart and ambitious).

We use word2vec (Mikolov et al. 2013), a well-adopted word vector representation technique, to quantify how likely women (relative to men) are to be linked to competence, as well as other traits (i.e., intelligence, warmth, masculine stereotypes, and feminine stereotypes), over time.[1]

Word2vec is a two-layer neural network which receives a corpus of text and transforms it to numerical vectors. This approach assigns each word a high-dimensional vector such that the relationship between vectors captures the semantic relationship between the words. Words that relate to one another such as fruit (e.g., apple and orange) or vehicles (e.g., car and truck) appear close together, but words that are not as related (e.g., apple and truck) appear further apart.

To position words, the approach considers words co-occurrence, distance, and occurrence in similar contexts. If "men are smart" shows up much more frequently than "women are smart," for example, that would increase the relative similarity (i.e., shrink the distance) between men and smart. The distance between occurrences of words also matters. Even though both "women are smart" and "women do great research and are also smart" contain the words women and smart, the first example has them closer together, which increases the similarity in word2vec space. Occurring in similar contexts also shapes similarity. If "women are scientists" and "smart people are scientists" both appear frequently, it increases the similarity between "women" and "smart" even if the two words never appear together. For example, as Figure 1 shows, words such as "he", "his", and "man" are located close to each other.

---

[1] We tested several other well-known methods on a random small subset of our data and found that word2vec was most appropriate. We compared truncated SVD (a.k.a. latent semantic allocation), Positive Pointwise Mutual Information (PPMI), GloVe (Pennington et al. 2014), and word2vec. The preliminary results and consistency of word2vec results along with previous research (Sahlgren and Lenci 2016) showed that word2vec using CBOW is the appropriate method for our question.

The similarity between word2vec vectors has been shown to be a powerful tool for studying language and human perceptions about the relationship between words. Bhatia (2017), for example, demonstrated that vectors are closer together for words that are similar or used in similar contexts. This can be seen as analogous to an implicit association test (IAT), where people demonstrate faster response time when asked to pair similar (than dissimilar) concepts (Caliskan, Bryson, and Narayanan 2017).

Consequently, recent work has begun to use similar methods to measure stereotypes. Work using the Google Books corpus (Lin et al. 2012), for example, found that word embedding associations (i.e., the similarity between male and female words and target words) captured human ratings of whether those target words were more associated with men or women ($r > 0.76$, $p < .001$; Kozlowski, Taddy, and Evans 2019). Other research using that corpus (Garg et al. 2018) found that word embedding associations between words related to women and different occupations words, tracked the percentage of women in each of those occupations over time ($r = 0.70$, $p < .001$). Consequently, a great deal of research in a variety of contexts has shown that word embeddings are a powerful method for capturing people's perceptions of the relationship between words, and those relationships over time.

To measure gender bias in music, we quantify the relationship between words related to each gender, and key dimension of interest (i.e., competence), over time. We used word lists from prior work (Garg et al. 2018) to identify words related to men (e.g., "man, "men," and "he") and women (e.g., "woman, "women," and "she," see Table A1 for full list). Then, we use word2vec to train a word embedding model per time-bucket and generate 300-dimensional real-valued word vectors. That means for each word (e.g., woman) and each period in the dataset (e.g., 1965-1970), we use the set of song lyrics over that period to position the vector for that word. Next, we take the word vector representation of words in our gender word lists and average the vectors to obtain one single 300-dimensional vector per each gender.

We use a similar approach to obtain vectors for key dimensions of interest (e.g., competence). In the case of competence, for example, we use a list of word from prior work (Nicolas, Bai, and Fiske 2021) that are known to relate to competence (e.g., smart, persistent, and knowledgeable, see Table A1 for more examples).

Then, to measure the association, we capture the similarity between gender vectors and key dimension vectors (e.g., competence) at any given period in time. We use cosine similarity metric because it is frequently used in the literature and in the original word2vec article (Mikolov et al. 2013). If a gender word and a competence word occur in similar contexts, for example, their vector representations are close in the word2vec space, which results in a larger cosine similarity score and indicates that the two words are highly associated. Conversely, if the words do not occur in similar contexts, the cosine similarity is smaller, and the words are considered less associated.

To give a sense of how this approach works, Figure 1 provides a simplified version. We only show two dimensions for the sake of illustration, but words are represented with 300-dimensional vectors in our analyses. As shown in the figure, semantically similar words (e.g., the group related to men or the group related to women) appear close to one another. As discussed, we create a single vector for each gender and compare how close it is to words representing key dimension of interest (e.g., the vector for the word "smart" which relates to competence). The figure shows that compared to the female vector, the male vector is closer to "smart," indicating that smart is more associated with men.

**Figure 1.** Simplified Illustration of Word Embeddings in Two-Dimensions.

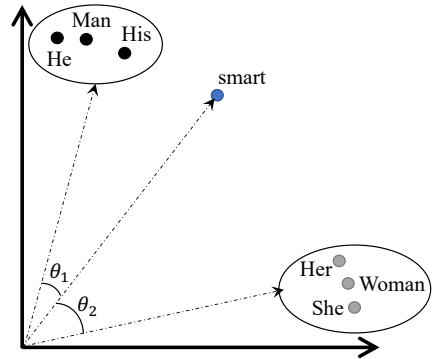

Note: Semantically similar words (e.g., the group related to women) appear close to one another. Because $\theta_1$ is less than $\theta_2$, cosine similarity of the male vector with the smart vector is larger than female vector with smart. This indicates that, compared to the female vector, the male vector is closer to "smart," suggesting that smart is more associated with men.

Formally, to calculate similarity of trait t (i.e., competence) to gender $g$, we calculate the cosine similarity of each word vector in $W_t$ to gender vector $V_g$, given that $W_t$ is a matrix of $n$ by 300 where $n$ is the number of words in trait $t$ (i.e., competence) and 300 is the vector size. To calculate how biased trait $t$ (i.e., competence) is, for each trait word (e.g., smart), we subtract cosine similarity between the trait word vector and the vector representing female from the cosine similarity between the trait word vector and the vector representing male. Positive values indicate that trait word is more associated with males, negative values indicate the opposite. This difference score also removes any general time trends that affect both genders equally (e.g., maybe more recent songs talk about competence more in general):

$$t \in \{competence\}$$

$$\overline{bias_t} = \frac{1}{n}\sum_{i=1}^{n}(cosine\_similarity(\overrightarrow{W_{ti}}, \overrightarrow{V_{male}}) - cosine\_similarity(\overrightarrow{W_{ti}}, \overrightarrow{V_{female}}))$$

In addition to prior work, further validation tests (see below) demonstrates that such difference scores accurately capture gender biases.

Finally, we analyze whether the associations change over time. We use linear mixed effect models with time as the independent variable and bias as the dependent variable. Given that association with gender may vary across words, we add a random effect for words. For ancillary analyses, we also run separate linear mixed effect models for each genre (e.g., rock).[2]

*Results*

Results suggest that while lyrics have become less misogynous over time, they remain biased (Figure 2). Words related to competence (e.g., smart) have become more associated with women, from strongly biased towards men to less so ($\beta$ = -.0002, p < .001). That said, gains level off, and potentially even reversed in the late 1990s ($\beta$ = -.0003, p < .001, $\beta^2$ = .00002, p < .001), and confidence intervals do not overlap with zero in recent years suggesting that lyrics still associate competence with men. Words related to intelligence (e.g., precocious and inquisitive, Garg et al., 2018) show similar results ($\beta$ = -0.40, p < 0.01, $\beta^2$ = -0.59, p < 0.001, Figure A3).

**Figure 2.** Misogyny in Lyrics Over Time.

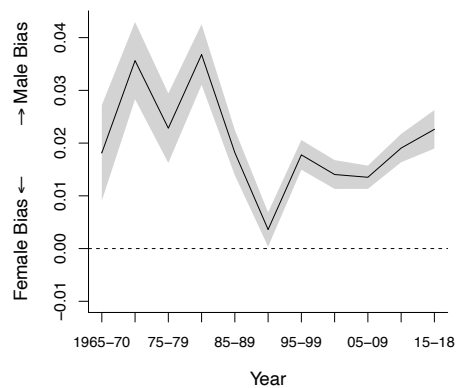

Note: Dashed line represents equal association with either gender and values greater (less) than 0 indicate the trait is more associated with men (women). Grey regions represent 95% confidence intervals around the estimates for each period.

Note that the leveling off is not driven solely by the introduction of new genres. Examining different genres separately (see Web Appendix) show that while the introduction of rap and dance in the 1980s contributed to the leveling off, even focusing on genres that existed previously (i.e., pop, rock, and country) shows similar patterns. More generally, ancillary analyses suggest that while competence has become more associated with women in pop, rock, and country over time, such associations have not changed in R&B and dance. In rap lyrics, competence has become less associated with women over time.

---

[2] We will release the codes and non-copy-righted data publicly upon acceptance.

For completeness, we also perform the same type of analysis for warmth related words (e.g., kind, friendly, and caring; (Nicolas et al. 2021, see Table A3 for more examples). Results indicate that warmth has become less uniquely associated with women ($\beta = 0.86$, p < 0.001, $\beta^2 = -0.53$, p < 0.001).

*Robustness Tests*

We also conducted several robustness checks to test whether alternative explanations can explain the results. We (1) used different word lists, (2) weighted genres differently (e.g., by their popularity), and (3) controlled for linguistic complexity (see Web Appendix). In all cases conclusions remained the same.

**Words used.** One could wonder whether the results might be driven by the particular words used. We test this a few ways.

First, as in previous research (Mikolov et al. 2013), we drop words with frequency smaller than 5 in each time bucket and repeat our analyses. The conclusions remain the same. Words related to competence become marginally less misogynistic ($\beta = -.0001$, p = .10) with a turnover in recent years ($\beta = -.0002$, p < .05, $\beta^2 = .00002$, p < .001).

Second, we only consider words which have occurred in every time point. This is a restricting rule to guarantee similar predictors (i.e., words) are compared over the years. Conclusions remain the same (linear: $\beta = -.0002$, p < .05; linear and quadratic: $\beta = -.0002$, p < .05, $\beta^2 = .00001$, p = .05).

**Songs used.** Alternatively, one could wonder whether the results are somehow driven by the genre weighting used. If the dataset happened to have twice as many rock songs as pop songs, for example, then rock music would have a disproportionate impact on the overall results. We address this in two ways. First, to ensure equal representation from each genre, our main analyses use under-sampling, a common machine learning practice when datasets include imbalanced class labels. Assume data from three genres $g_1$, $g_2$, and $g_3$, where $g_1$ has the smallest number of songs. To generate a balanced sample for the cross-genre analyses, we randomly select $||g_1||$ songs from genres $g_2$ and $g_3$ and perform the analyses on that sample. We repeat this sampling and analysis 100 times, averaging across the runs. Substantial number of dance and rap songs do not enter the data until 1980 and 1990 respectively, so we only include them (and sample accordingly) after those time points.

Second, one could argue that certain genres are less popular and so equal weighting does not actually reflect reality. To test this possibility, we re-run the model weighting the genres based on popularity. Billboard Hot 100 charts are the only consistent audience consumption data available from

1960s to today, so we rely on genre popularity scores collected from Billboard by Beckwith (http://thedataface.com/2016/09/culture/genre-lifecycles), indicating the percentage of songs in Hot 100 charts from each genre in each time period. We weight things accordingly. In 2000, for example, imagine country, rock, and pop songs occupy 25%, 25%, and 50% of the Billboard charts respectively. If there are 1000 songs from each genre in the dataset, we select 1000 pop songs, and 500 songs from rock and country genres (total of 2000 songs). The number of songs for 1965-69 time bucket did not meet our threshold of half a million words and was dropped for the genre popularity analysis.

Conclusions remain the same. Words related to competence ($\beta$ = -.0003, p < .001) become less associated with men over time, though things have slightly reversed as of late ($\beta$ = -.0003, p < .001, $\beta^2$ = .00002, p < .001).[3]

*Potential Consequences*

One might wonder whether these lyrical shifts have any consequences. Might they change how the population at large sees men and women, for example, or do they simply reflect changes that have already occurred? Said another way, do shifts in lyrics *precede* shifts in societal stereotypes (suggesting they cause or are a leading indicator of societal change), or do they tend to *follow* them, indicating that music merely reflects what has already occurred in society at large?

To examine these possibilities, we leverage prior work aggregating all available public opinion polls from 1946 to 2018 measuring gender stereotypes related to competence (Eagly et al. 2019). These polls asked respondents whether certain characteristics (e.g., competence) was more true of women, men, or equally true of both and the number of respondents that choose women, divided by the number that selected either women or men, captures gender stereotypes over time (Eagly et al. 2019).

Results indicate that misogyny in lyrics is strongly correlated with societal stereotypes (r = .62). Importantly, this relationship does not simply reflect two time series moving in the same direction. Lyrical changes are most strongly related to *subsequent* public opinion (i.e., opinion polls that directly follow the time period of lyrics). Examining lyrics and *simultaneous* public opinion (i.e., opinion polls that occur in the same time period as the lyrics) or *preceding* public opinion (i.e., opinion polls that appear in the time period before the lyrics) show weaker or reversed relationships (i.e., r = .30 and r = -.64 respectively). These results suggest that changes in lyrics may contribute to, or be a leading indicator of, shifts in stereotypes held by the population at large.

---

[3] Alternatively, one could wonder whether, given that more popular songs may be more likely to have their lyrics available, the dataset, especially in earlier years, may include more popular songs. That said, because more popular songs are listened to more anyway, they should have a greater impact on the population.

*Drivers of Lyrical Shifts*

One might wonder what is driving the observed changes. Are the linguistic shifts similar for male and female artists, for example? Or might they simply reflect an increase in the number of female artists, who may use less misogynistic and gendered language?

To begin to address these questions, we analyze artist gender. To do so, we rely on methods used in prior research (Santamaría and Mihaljević 2018; Squire 2019) to find artists' gender. To infer artists' gender, we use gender_guesser package (version 0.4.0) in Python along with national name list (Kaggle 2017). For individual artists (e.g., Katy Perry or Billy Joel) we simply use their name, but for artists that go by pseudonyms (e.g., Snoop Dogg) or bands that have multiple members (e.g., Metallica or the Supremes) we first use their Wikipedia page to extract the individual name(s). In cases where no such page is available, research assistants searched for this information manually. Then, we use a similar approach as individual artists to detect gender and assign the band a score based on the percentage of members that are female. Gender was able to be assigned for 99.5% of the artists.[4]

Results cast doubt on the notion that the decrease in misogyny is driven by the introduction of more female artists. There is no significant change in the percentage of female artists over time ($\beta$ = -.0001, p = .82). Further, while one could argue that this is driven by new genres emerging, even within genres there is no significant change ($\beta_{pop}$ = .0037, p = .06, all other genres p > .2). This casts doubt on the possibility that shifts in language are driven by more female artists over time.

Instead, results are more consistent with a shift in male artists' language (Figure 3). For male artists, words related to competence ($\beta$ = -.0004, p < .001) become more associated with women. While female artists' language shows less evidence of change ($\beta$ = .0001, p = .35), this is likely due to the fact that female artists' language was less misogynistic to begin with (Figure 3). Female artists are more likely to associate women with competence overall ($\beta$ = -.0144, p < .001), and the gender x time interaction ($\beta$ = .0002, p < .001) indicates that the gender difference has decreased over time as male artists started associating women more with competence. Analysis of the most recent time period, however, suggests that artists of both genders remain biased. Lyrics associate men more with competence.

---

[4] Note that there are three to four times more male artists in the dataset, so the effects for male language may be more precisely estimated. Further, while there are enough solo female artists or all female groups, and solo male artists or all male groups to estimate effects for each, it is more difficult to do so for mixed gender bands. Less than 10% of songs were performed by mixed gender groups which provides limited data for estimating language changes.

**Figure 3.** Misogyny in Lyrics Over Time for Male and Female Music Artists.

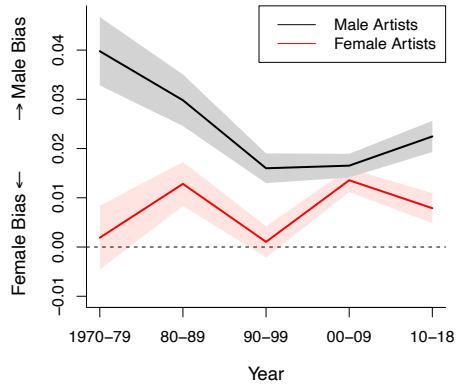

Note: Dashed line represents equal association with either gender and values greater (less) than 0 indicate the trait is more associated with men (women). Grey regions represent 95% confidence intervals around the estimates for each period.

*Validation*

As noted, research demonstrates that word embeddings can be used to accurately capture gender biases (DeFranza et al. 2020; Garg et al. 2018; Kozlowski et al. 2019; Lewis and Lupyan 2020). That said, to ensure that word embeddings are capturing misogyny and gender stereotypes in our context, we conducted additional validation checks.

First, we test whether the embedding bias tracks Kozlowski et al.'s (2019) manual coding. They asked Amazon Mechanical Turkers to rate 59 words (e.g., "nurse") based on how much they are associated with a gender (0 = very feminine, 100 = very masculine). The embedding bias scores derived from our approach at the most recent time point (i.e., 2015-2018) are highly correlated with their human-coded scores ($r = .63, p < .001$), supporting the notion that the embedding bias captures the gendered association of different words.

Second, we perform a similar test on competence words used in our analyses. Three research assistants coded 286 competence words (e.g., think and knowledge) based on how stereotypically masculine or feminine they were (i.e., -2 is very feminine word, 2 is very masculine word). As expected, results demonstrate that the embedding bias at the most recent time point (i.e., 2015-2018) is reasonably correlated with human-coded stereotype scores ($r = .41, p < .001$), further supporting the notion that the embedding bias captures the gendered association of different words.

Third, while the lack of historical human ratings makes it difficult to validate the embedding bias in prior time periods, following Garg et al. (2018), we leveraged human-coded gender stereotype scores at two time points, 1977 and 1990 (Williams and Best 1977, 1990). This data involved participants rating 300 adjectives (e.g., "headstrong" and "talkative") based on words association with men or women. Since we use 5-year time buckets, we trained models for 1975-1979 and 1990-1994 periods respectively. We then compared 1977 human-coded scores to 1975-1979 model and 1990 human-coded scores to 1990-1994 model. As expected, the word embedding bias was significantly correlated with human-coded gender stereotype scores for both time periods ($r = .36, p < .001$ and $r = .18, p < .05$), underscoring the notion that word embeddings are accurately capturing differences in gender associations.

Finally, we also provide a more general test that our word embeddings are accurately capturing the semantic relationship between words. As mentioned earlier, words that are similar or are used in similar contexts, have been shown to have closer word vectors in the word embeddings space (Bhatia 2017). We find the same thing in our data. Following Hamilton, Leskovec, and Jurafsky (2016), we use a standard similarity task benchmark called MEN (Bruni et al. 2012) which includes human similarity judgments of 3000-word pairs. As expected, the similarity score between these word pairs in our model is highly correlated with the pair-wise human-coded similarity ($r = .47, p < .001$). This further underscores that word associations in our trained model well represent human perceptions.

**GENERAL DISCUSSION**

Researchers and cultural critics alike have long debated whether popular music is misogynous. But questions of scale (e.g., examining only one genre over a few years) and measurement have limited the ability to draw strong conclusions. Consequently, it remains unclear whether lyrics are misogynous, whether this has shifted across time, what may have driven any shifts, and how such shifts relate to public opinion.

Analyzing over a quarter of a million songs over more than 50 years sheds light on these questions. First, while there is little evidence of explicit misogyny (e.g., greater aggression towards women), subtler forms of misogyny suggest a more complex picture. Lyrics have become less biased against women over time, but bias remains. Women are more likely to be associated with competence and intelligence than they were in the past, but these concepts are still more associated with men. Lyrics have become less gendered more broadly, but they remain gendered. Further, reductions in bias seem to have slowed and may even have reversed in some cases. The results not only provide empirical evidence, but allow for quantitative comparison of where (i.e., which genres and artists) and when (i.e., time periods) biases may be larger.

Second, ancillary analyses suggest potential consequences of lyrical changes. Lyrical shifts are strongly correlated with subsequent changes in public perceptions. These results suggest that if lyrics are not driving broader cultural changes, they are at least a leading indicator.

Third, rather than being driven by more female artists, these results are driven by male artists shifting their language over time. Further, they are driven by both the introduction of new genres (e.g., rap) and changes in older ones (e.g., country).

*Broader Applications*

While our empirical work focused on misogyny in music, word embeddings can be used to shed light on a variety of topics. This includes questions related to racism and discrimination, and well as consumer research more broadly.

To study discrimination, for example, researchers could survey consumers, asking about how they perceive brands communicate with different audiences. In cases where discrimination and bias are subtle, however, they may be difficult to capture. Further while this works reasonably well when looking at one, or couple points in time, measuring their stance over a longer time horizon becomes more challenging. Collecting survey data over decades is often prohibitive, and even if prior researchers collected some related data, they often used different measures.

Word embeddings can help address these challenges. Rather than asking people for their perceptions, researchers can use existing textual data (e.g., social media or online reviews) to automatically measure semantic relationships. Just as we measured the relationship between words related to women and competence, researchers could examine whether brand use different language when they target different racial groups or genders, and whether this changes over time. Similar approaches could be used to track how brands talk about diversity, equity, and inclusion and how it shifts over time.

Beyond racism and discrimination, similar approaches can be used to study brand perceptions, goals, self-regulation and a range of other topics. Work on desire and desire regulation (Hofmann et al. 2012), for example, notes that "the majority of research on self-regulation occurs in the laboratory" and that "little is known about what types of urges are felt strongly (or only weakly), what urges conflict with other goals, and how successfully people resist their urges" (p. 582). Word embeddings, topic modeling, and similar approaches can allow researchers to examine these questions in more naturalistic settings. Social media, blog posts, and even the Google Books dataset could be used to examine the urges people have, how they conflict, and even how these urges may have varied over time. What people want and feel like they should do today (e.g., balance work and family or use less social media), for example, is likely very different today than it was 20 years ago.

*Conclusion*

In conclusion, word embeddings and other computational linguistic techniques provide a powerful toolkit to study culture. Researchers have long been interested in quantifying cultural dynamics, but measurement has been a key challenge. Natural language processing, however, provides a reliable method of extracting features, and doing so at scale. These emerging approaches have the potential to shed light on a range of interesting questions, both in cultural analytics, and more broadly.

# WEB APPENDIX

## Are Women More Likely to Be the Object of Aggression?

Given prior work often measures misogyny using aggression towards women, we test this as an ancillary analysis. We measure the degree to which women and men are the recipient of aggressive thoughts and actions. In phrases like "I hit her", for example, "her" is the recipient (i.e., object) of the aggressive verb "hit." To identify verbs commonly used for expressing aggression (e.g., hate and burn) we use Violence Vocabulary Word List (Anon 2019). We use the spacy package (version 2.0.18) in Python to extract words dependency in sentences and calculate the frequency of a gender being used as an object of an aggressive verb.

Both genders show increasing trend in terms of being recipients of aggression, but their relative ratio does not change. While women are more likely to be recipients of aggression in songs today than in the 1960s ($\beta$ = 1.45, p < .001), this ignores the possibility that aggressive acts could have generally increased over time and affected both genders equally. Indeed, men are also more likely to be recipients of aggression in recent years ($\beta$ = 1.70, p = .07). Looking over time shows that women (47.6%) and men (52.4%) are recipients of aggression a similar amount overall ($\chi^2(1)$ = 2.50, p = .11) and that this does not change significantly over the years ($\beta$ = -.0026, p = .86).

One could wonder whether the results are somehow driven by the specific way in which the outcome measure was calculated, so to test robustness, we also use an alternate approach. We calculate the difference between women and men being recipient of aggression while controlling for how often they are recipient of *any* thought or action (e.g., "I kissed him" or "I kissed her"). Consistent with the main analysis, however, this difference between the ratios does not change over time ($\beta$ = -.0001, p = .77).

**Table A1.** Word lists used to capture gender and competence.

| Female | she, daughter, hers, her, mother, woman, girl, herself, female, sister, daughters, mothers, women, girls, females, sisters, aunt, aunts, niece, nieces |
|---|---|
| Male | he, son, his, him, father, man, boy, himself, male, brother, sons, fathers, men, boys, males, brothers, uncle, uncles, nephew, nephews |
| Competence (examples) | able, ambitious brave, determined, fighter, knowledgeable, master, smart, strong, talented, reasonable, realistic, persistent |

## Within Genre Analysis

Ancillary analyses suggest that the cross-genre effects may be driven by the introduction of new genres as well as changes in older genres. Averaging just pop, rock, and country together, suggests that these genres have become less misogynistic ($\beta = -.0004$, $p < .001$), but there is only a marginal curvilinear effect ($\beta = -.0005$, $p < .001$, $\beta^2 = .00009$, $p = .06$; Figure A1). The same trend holds for intelligence (linear: $\beta = -.0020$, $p < .001$; linear and quadratic: $\beta = -.0023$, $p < .001$, $\beta^2 = .00006$, $p = .13$), warmth (linear: $\beta = .0005$, $p < .001$; linear and quadratic: $\beta = .0006$, $p < .001$, $\beta^2 = -.000003$, $p = .47$), masculine stereotypes (linear: $\beta = -.0014$, $p < .01$; linear and quadratic: $\beta = -.0016$, $p < .001$, $\beta^2 = .00005$, $p = .08$), and feminine stereotypes (linear: $\beta = .0010$, $p < .05$; linear and quadratic: $\beta = .0010$, $p < .05$, $\beta^2 = -.00002$, $p = .54$).

**Figure A1.** Gender association with competence for pop, rock, and country combined.

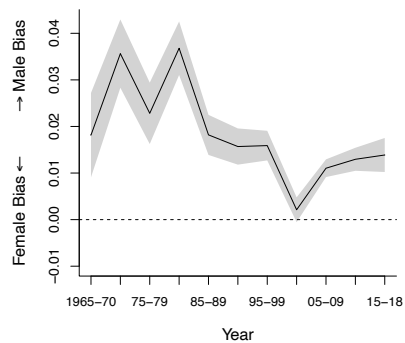

Note: Dashed line represents equal association with each gender and values greater (less) than 0 indicate the trait is more associated with men (women). Grey regions represent 95% confidence intervals around the estimates for each period.

These overall trends, however, obscure some genre specific variation (Figure A2, Table A2). In pop, competence has become more associated with women over time ($\beta = -.0009$, $p < .001$) and is actually slightly more associated with women (than men) today. Similar results hold for rock and country ($\beta_{rock} = -.0005$, $p < .001$; $\beta_{country} = -.0004$, $p < .05$) though competence remains more associated with men. R&B and dance show no change ($\beta_{R\&B} = .0001$, $p = .70$; $\beta_{dance} = -.0002$, $p = .36$). In Rap, competence has become less associated with women ($\beta = .0015$, $p < .001$). Note, however, that women were relatively more associated with competence in rap lyrics than other genres to begin with. Thus, while woman have become less associated with competence in rap lyrics over time, this may be partially due to the fact that they were more associated with women initially.

**Figure A2.** Within genre gender association with competence.

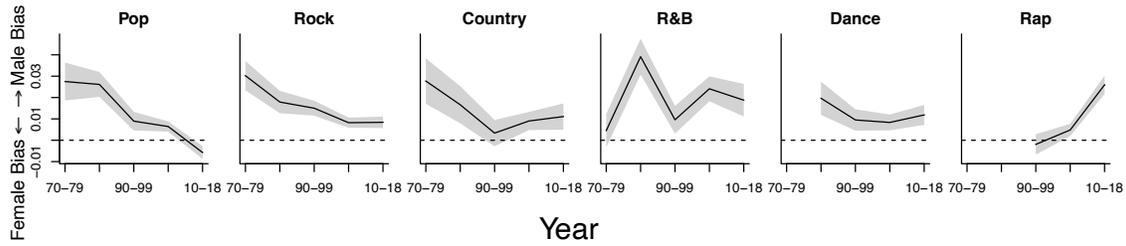

Note: Dashed line represents equal association with each gender and values greater (less) than 0 indicate the trait is more associated with men (women). Grey regions represent 95% confidence intervals around the estimates for each period.

**Table A2.** Within genre gender association with competence.

| Genre | $\beta$ | $\beta^2$ |
|---|---|---|
| Pop | -.0009 *** | -.00000 |
| Rock | -.0005 *** | .00001 ^ |
| Country | -.0004 * | .00003 * |
| R&B | .0001 | -.00001 |
| Dance | -.0002 | .00003 |
| Rap | .0015 *** | .00008 ^ |

*Note*. *** p < .001, ** p < .01, * p < .05, ^ p < .10

## Ancillary Analyses

We also ran several ancillary analyses that, taken together, suggest that lyrics have become less gendered more broadly.

We examine words related to intelligence more specifically (e.g., analytical; (Gaucher, Friesen, and Kay 2011). Results remain the same. Words related to intelligence (e.g., thoughtful) have become more associated with women, from strongly biased towards men to less so ($\beta$ = -.0009, p = 0.09). That said, gains level off, and potentially even reverse in the late 1990s ($\beta$ = -.0016, p < .01, $\beta^2$ = .0002, p < .001), and confidence intervals do not overlap with zero in recent years suggesting that lyrics still associate competence with men (Figure A3).

**Figure A3.** Gender association with intelligence.

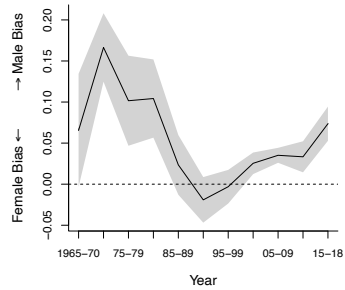

Note: Dashed line represents equal association with each gender and values greater (less) than 0 indicate the trait is more associated with men (women). Grey regions represent 95% confidence intervals around the estimates for each period.

For completeness, we examine the association between gender and warmth. To do so, we use the same word embedding approach, but calculate word vectors for warmth related words each time period (e.g., words like "kind" and "charming", Table A3, (Nicolas, Bai, and Fiske 2021). Results indicate that warmth has become less associated with women ($\beta$ = .0003, p < .001), though these changes have leveled off ($\beta$ = .0003, p < .001, $\beta^2$ = -.00001, p < .001).

Further analysis of words associated with masculine stereotypes (e.g., leader and ambitious) and feminine stereotypes (e.g., cooperate and trustworthy (Gaucher et al. 2011)) suggests that lyrics may have become somewhat less gendered overall.[5] Words related to masculine stereotypes have become less associated with men ($\beta$ = -.0012, p < .01) and words related to feminine stereotypes have become less associated with women ($\beta$ = .0011, p < .05). For masculine stereotypes, though, gains level off in the early 1990s ($\beta$ = -.0018, p < .001, $\beta^2$ = .00009, p < .01).

## Robustness Tests

### Complexity

One could also wonder whether the results simply reflect the increasing complexity of song lyrics. This argument would suggest that while song lyrics in the 1970s were rather simplistic (e.g., "let it be"), today's lyrics may be much more complex and varied and this would somehow lead to our effect.

But while such an argument might suggest that the increased complexity would lead to an increased distance between men and competence words, as well as women and competence words, it says less about why the *relative* distance between women and men with competence words would decrease.

---

[5] One might wonder whether lyrics becoming less gendered (i.e., less associated with either gender) is good. While decreased gendered associations is clearly good in some situations (e.g., competence moving from being more associated with men to less so), in others it is less obviously beneficial (e.g., warmth becoming less associated with women). Thus, whether it is perceived as good or bad seems to depend less on the change and more about whether the trait is seen as positive and originally associated with women or not. Consequently, we use the termed gendered rather than biased to be agnostic about the valence of the change.

Said another way, it would suggest that the distance between both male and female words and competence would increase, but complexity alone has difficulty explaining why the increase would be greater for one gender than the other.

Further, results persist when controlling for language complexity over time. To control for the possibility that lyrics may have become more complex, we measure how the distance amongst words within each of the gender word lists (e.g., words related to men like "man, "he," and "his") has changed over time. This captures whether the similarity of even related words ("man" and "he," for example, or "woman" and "she") has shifted over time. We then take the average of the similarity among words related to men (e.g., "man, "he," and "his") and similarity among words related to women (e.g., "woman, "she," and "her") and normalize our main competence measure by this average. The conclusions remain the same. Competence words have become more associated with women over time, though they remain more associated with men. Further, these gains have leveled off and even reversed as of late.

**Table A3.** Word lists used to capture warmth, intelligence, feminine and masculine stereotypes.

| Warmth (examples) | attached, caring, charming, faithful, forgiving, friendly, honest, humble, innocent, lovely, peaceful, sincere, sympathetic, tender, warm. For full competence and warmth dictionaries see Nicolas et al. (Nicolas et al. 2021). |
|---|---|
| Intelligence | precocious, resourceful, inquisitive, sagacious, inventive, astute, adaptable, reflective, discerning, intuitive, inquiring, judicious, analytical, luminous, venerable, imaginative, shrewd, thoughtful, sage, smart, ingenious, clever, brilliant, logical, intelligent, apt, genius, wise |
| Feminine Stereotypes | affectionate, cheerful, cheery, cheerfully, childish, childishly, committed, communal, compassionate, connected, considerate, cooperative, dependent, empathetic, emotional, feminine, flatterable, gentle, honest, interdependent, interpersonal, kind, loyal, modest, nag, nurture, nurturer, nurturing, pleasant, polite, quiet, responsive, sensitive, submissive, supportive, sympathetic, tender, trusting, understanding, warm, whiny, whining, whiny, yielding, whine |
| Masculine Stereotypes | active, adventurous, aggressive, ambition, ambitious, analytical, assertive, athletic, athletically, autonomous, boastful, boasting, challenges, competitive, competitor, confident, courage, courageous, decisive, determination, determined, dominant, forceful, forcefully, greedy, headstrong, hierarchical, hostile, impulsive, independent, individual, individuality, intellectual, leader, leadership, logical, masculine, objective, opinionated, outspoken, persistent, persists, principled, reckless, self-confident, self-reliant, self-sufficient, stubborn, superior |

We also considered alternate analyses beyond word embeddings. First, we considered testing whether there is a shift in how often men and women appear as subjects and objects. The subject of a sentence has an active role as the agent of action (e.g., the word she in "she called him"), while the object takes a more passive role as the recipient of action (e.g., the word him in "she called him"). We could thus measure how often men and women appear as subjects (e.g., she and he) and objects (e.g., her and him) over time. The link between this and misogyny, however, is not completely straightforward. Women could be the subject of a sentence, for example, but still talked about negatively (e.g., she is dumb). Second, we tried to examine changes in derogatory words (e.g., b*tch or h*) over time, but measurement proved challenging. For example, while "b*tch" was originally referred to women, its meaning shifted to be used for both genders. Compared to "smart" which has had consistent meaning from 1960s to today. Thus, it is difficult to accurately capture shifts in reference to women.